\documentclass[a4paper]{article}
\usepackage[margin=2.5cm]{geometry} 
\usepackage{threeparttable}
\usepackage{tabularx}
\usepackage{soul}
\usepackage{graphicx, color}
\usepackage{float}

\usepackage{url}
\usepackage{CJKutf8}
\usepackage[unicode, breaklinks, urlcolor=cyan, linkcolor=blue]{hyperref}

\usepackage{amsmath}
\usepackage{amsthm}
\usepackage{amssymb}

\usepackage[authoryear, comma, sort]{natbib}

\title{Analyst Reports and Stock Performance: Evidence from the Chinese Market}

\author{%
  Rui LIU$^{1}$,
  Jiayou LIANG$^{1}$,
  Haolong CHEN$^{1}$,
  \and
  Yujia HU$^{1}$\thanks{Corresponding Author: \texttt{yujiahu@uic.edu.cn} }\\
  \vspace{0.5cm}
  \small $^{1}$ Faculty of Science and Technology, BNU-HKBU United International College, Zhuhai, China \\
}

\date{ April 11, 2024}

\begin{document}

\maketitle

\begin{abstract}
This article applies natural language processing (NLP) to extract and quantify textual information to predict stock performance. Using an extensive dataset of Chinese analyst reports and employing a customized BERT deep learning model for Chinese text, this study categorizes the sentiment of the reports as positive, neutral, or negative. The findings underscore the predictive capacity of this sentiment indicator for stock volatility, excess returns, and trading volume. Specifically, analyst reports with strong positive sentiment will increase excess return and intraday volatility, and vice versa, reports with strong negative sentiment also increase volatility and trading volume, but decrease future excess return. The magnitude of this effect is greater for positive sentiment reports than for negative sentiment reports. This article contributes to the empirical literature on sentiment analysis and the response of the stock market to news in the Chinese stock market.



\end{abstract}

\section{Introduction}
\label{sec: intro}

In the era of information, text plays a crucial role in shaping financial market behavior, as it serves as a medium of communication, conveying thoughts, emotions, and experiences. Investors and analysts rely on various text sources, including news, social media, and professional reports, to gather information and make informed financial decisions. As a result, the impact of textual information on the financial sector has gained significant attention, leading to the emergence of a cross-disciplinary research field that combines Natural Language Processing (NLP) with financial analysis for the evaluation of securities.

The analysis of text reports for investment decision making poses several challenges. First, these reports are typically written objectively and unemotionally, making it difficult to extract sentiment and provide clear investment advice. Overcoming this challenge requires finding effective methods to accurately identify the emotions conveyed in reports. Second, research reports could have a bias toward presenting a positive outlook for companies, particularly on the Chinese market due to incentives and market conditions (see \cite{wu2018optimistic}). Therefore, distinguishing a real positive sentiment or a rather negative sentiment from a large number of reports becomes difficult, as it necessitates information on the analyst to control for her incentives. Lastly, evaluating the validity of information in research reports is another challenge, considering that data can be sensitive to time and quickly become outdated. To make informed investment decisions, investors need ways to assess the accuracy and relevance of the data in the report articles.

In light of these challenges, this study aims to quantitatively extract sentiment from investment recommendation reports on the Chinese stock market and examine the impact of these sentiments on stock performance. Specifically, we investigate whether the sentiments expressed in the reports influence future excess returns, volatility, and trading volume. To achieve this objective, we employ the BERT (Bidirectional Encoder Representations from Transformers ) language model to analyze sentiment in a dataset comprising $62735$ Chinese financial analyst reports. Then we establish a relationship between the sentiment expressed in the reports and the behavior of the stocks with regression analysis. Our findings indicate several notable insights. First, we observe a positive relationship between the sentiment scores of the research reports and the excess returns of the stocks the next day. Additionally, strong sentiment, both negative and positive, leads to increased intraday volatility of stocks the following day. Furthermore, we find that strong positive sentiment is associated with higher trading volume on the next day, whereas the same effect is not necessarily observed for negative sentiment.

Our work uses a new dataset of collected analysts' reports that, to the best of our knowledge, have never been collected and used before in scholarly research. Similarly as in other research work, it uses a deep learning model to extract sentiment from text. Our contribution consists in providing empirical evidence of sentiment effects arising from analyst reports and impacting stock market performance in China. Although the impact of implied sentiment measures derived from different methodologies has been applied to study its effect on the Chinese financial market such as in \cite{wu2018optimistic} or in \cite{liang2020sentiment}, this work is the first, to our knowledge, that uses a language model to train and predict sentiment in a text corpus and investigate its effect on the performance of the Chinese financial market. The impact of direct investment recommendation by Chinese analyst has been studied \cite{jiang2014information} finding significant investment performance effects of these recommendations. Our work uses a more extensive dataset and provides additional empirical evidence of information content of analysts' reports.

Our findings shed light on the importance of sentiment analysis in the financial domain and provide valuable implications for investment decision-making. The remaining structure of the article is as follows. Section \ref{sec: lit_rev} reviews the research literature on sentiment analysis and NLP application in finance. Section \ref{sec: methodology} introduces our methodology for the analysis of sentiment and regression models. Section \ref{sec: results} contains the results of the analysis. Finally, Section \ref{sec: conclusion} concludes our work.


\section{Literature Review} 
\label{sec: lit_rev}

With the emergence of social networks and the availability of social media data, a strand of research studies the impact of the publication of social media text on market performance. Initially, they focused on simple, easily accessible news and commentary from social networks. For example, \cite{tetlock2007giving} uses factor and regression analysis on data from the General Inquirer program and Dow Jones Newswires to construct a "pessimism factor" from media content and concluded that high levels of media pessimism predict temporary decreases in market returns and increased trading volumes, suggesting a link between media content and investor sentiment. \cite{bollen2011twitter} use mood tracking tools, named OpinionFinder and Google-Profile of Mood States, to analyze whether the text content of the daily Twitter feeds is correlated with the value of the Dow Jones Industrial Average (DJIA) and find significant predictive results. More recently, \cite{zeitun2023impact} reveal that Twitter-based sentiment influences the returns of various US sectors differently and the aggregate S\&P 500, and \cite{XU2023100757} investigate the impact of the COVID-19 pandemic in the United States on stock market reactions to market and individual company news and find that the pandemic causes investors to overly focus on macro news while neglecting micro details, leading to increased stock reactions to market news and subdued responses to firm-specific news during the pandemic. \cite{audrino2020impact} find predictability of other performance measures such as stock volatility due to the sentiment effect by conducting experiments with an extensive data set of financial news. 


 Although these opinionated texts and short news can impact the market due to the uneven quality of the information they contain, research has begun to shift toward more professional reports. In a review, \cite{brauer2018analyzing} argued that financial analysts with professional knowledge can present market information in a more concentrated form through reports and impact equity in more significant ways. \cite{huang2018analyst} believe that analysts play the role of information intermediary by discovering information beyond corporate news releases. Such textual information is often more valuable and can better guide investment companies in responding to market changes.  Therefore, people have started studying the impact of analyst behavior and the content of their reports on the market. For example, \cite{asquith2005information} investigates the association between market returns and the content of security analyst reports, and the results suggest that the market reaction is influenced by the strength of the arguments in the analyst report and the affiliation of the analyst’s brokerage firm. 


Other studies focus on the impact of market sentiment instead of specific sentiment in financial text. In these studies, \cite{baker2006investor} use six different features to construct an investor sentiment index (BW index) and suggest that when sentiment is estimated to be high, stocks tend to earn relatively low subsequent returns. \cite{stambaugh2012short} apply the BW index and find that there are market-wide variations in investor sentiment, such as overpricing that can occur for specific stocks during periods of high sentiment. Based on the BW index, \cite{huang2015investor} develop an aligned sentiment index and their findings suggest that high sentiment causes overvaluation of the aggregate stock market, leading to a low future market return. \cite{ASHOUR2023100755} also use indices obtained as the principal component of other economic indicators. In these studies, market sentiment is a predominantly highly significant factor in stock performance. Our article is more related to the previous group in the sense that we analyze the sentiment content of financial analyst recommendation reports to the performance of individual stocks.

In the area of accounting and financial statement analysis, NLP methods have been widely studied. \cite{r5} conducted a review of NLP applications in financial market research.  They identified corporate governance, accounting report verification, and risk management as the main areas where NLP has been applied. Many studies focus on the effects of earnings announcements, such as \cite{beaver2018information} who find that the information content of earnings announcement reports has a variety of impacts to different types of companies and \cite{ARDIA2022100683} who conduct a tone-based event study and find that the abnormal tone in media articles helps explain price reactions around an earnings announcement and predicts subsequent price dynamics in the 20 days following the event, and they conclude that the media provides additional information not included in the information from earnings press releases and earnings calls.
 
Most of the related empirical literature reports significant investor reactions to analyst reports. For example, \cite{howe2009predictive} uses more than 350,000 analyst recommendations to examine the predictive content of the aggregate analyst recommendations and finds that they contain market- and industry-level information about future returns and earnings. \cite{asquith2005information} analyzed a sample of 1,126 reports written by 56 different analysts. The study included variables such as the percentage change in the analyst's projected earnings forecast and the price target for a particular firm at a given time. The study also looked at information announcements, such as earnings announcements, dividend changes, and management changes, that occurred within a certain time frame around the release of the analyst report.  Based on 363,952 analyst reports, \cite{huang2014evidence} find that investors react more strongly to negative than positive texts, suggesting that analysts are important in declaring negative information.  On the other hand, \cite{loh2010investor} finds that investors do not react to stock recommendations and suggests investor inattention as the reason.

Most of the research discussed so far uses financial data and reports from the US market. In the analysis of investors reactions to analyst reports there are regional differences based on the market environment.  For example, \cite{Hsieh} examined how different types of investors in the Taiwan market react to information from securities analyst reports. Foreign institutions and domestic mutual funds are the primary users of analyst reports, and their buy-sell imbalances move in tandem with analyst signals and significantly explain the size of cumulative abnormal returns across incidents of analyst report releases. For the Australian market, \cite{Kim} investigate the role of analysts in the information environment of the market. They find that the presence of a larger number of analysts covering a firm is associated with a lower level of information asymmetry and a higher level of price informativeness. 

In mainland China, research on text analysis and the impact of those on the financial market is not very widespread. \cite{huang2020textual} review research in this area and conclude that this is still in a starting stage. \cite{huang2021reflects}  formulate different types of sentiment through text data, stock prices, and model estimates of financial volatility, and they find that the text-based sentiment index has the strongest correlation with the stock market. \cite{jiang2014information} study abnormal performance in the Chinese stock market from analyst recommendation revisions, and, similarly, \cite{jia2017market} study the reaction of prices to analyst recommendations in the mainland and Hong Kong markets, and find a different price sensitivity to local and foreign analysts. However, both articles do not use a language model to extract the analyst's recommendation sentiment. In terms of methodology and data, the paper of \cite{Liang} is the most similar to ours, since it applies a deep learning model (LSTM) to extract textual tones from analyst reports written in Chinese. The research employs scraped website data that, unlike the data in our study, are likely not written by professional financial analysts. Furthermore, our study uses a state-of-the-art deep learning architecture to train sentiment from text data.

In relation to the model used for natural language processing, the BERT (Bidirectional Encoder Representations from Transformers) is currently the most popular model for classification of text articles. The model architecture was first proposed by a team of Google researchers in the paper \cite{devlin2018bert}. The model has been pre-trained on a large amount of text data. It can understand complex patterns in language and can be used for various NLP tasks such as sentiment analysis, entity recognition, and question-answering systems, among others. Recently, this and subsequent BERT models have been used in financial market research. For example, \cite{man2020stock} apply the BERT model to conduct sentiment analysis in financial news. Their results can guide stock trading based on positive, negative and neutral sentiments. \cite{li2021sentiment} also adopted the BERT for sentiment analysis model in Chinese stock reviews. They focus on a different architecture based on fully connected layers and find reliable classification accuracy.


\section{Methodology}
\label{sec: methodology}

\subsection{BERT Model Training}
\label{subsec: BERT}

The language model we use is the Chinese pre-trained BERT model developed jointly by Harbin Institute of Technology (HIT) and Microsoft Research Asia (MSRA) described in \cite{cui-etal-2021-pretrain}. It is officially called Chinese BERT with Whole-Word Masking, abbreviated as ``Chinese BERT-wwm". This model is based on the original BERT model (\cite{devlin2018bert}) and adopts the full-word masking strategy during pre-training. Unlike the original BERT model, Chinese BERT-wwm masks the entire word as a whole during pre-training, instead of randomly masking a character within the word. This approach is more suitable for the characteristics of the Chinese language and allows better capturing of semantic information in Chinese words.

The ``Chinese BERT-wwm" with its original architecture, weights, and vocabulary is fine-tuned with our data set of financial analyst reports for sentiment classification. The final prediction layer is a softmax function:
\begin{equation}
    sent_{i} = \textbf{softmax} (hW+b)    
\end{equation}
where $sent_i$ is the sentiment for a text report $i$ and has three classes: positive, neutral or negative. The linear transformation $hW+b$ is a projection of the vocabulary space $h$ with weights $W$ and bias $b$ that are fine-tuned. This fine-tuning process is computationally relatively inexpensive, as it requires a few hours on GPU.

For the fine-tuning process, we need labeled training data. Because of challenges and subjectivity in labeling sentiment discussed previously, we adopt an automated labeling process based on the industrial excess return for each stock in a 3 days horizon around the report release date. Specifically, for each report, we identify the stock (or stocks if the report is related to multiple stocks) and the date this report is released. Then, we compute the industrial excess return of the stock in 3 days (a day before the release, the day of the release and the day after the release) and compute their average excess return:
\begin{equation}
R^{ex}_{i,t} = \frac{1}{3} \sum_{t=-1}^{1}\left( r_{i,t} - \overline{r_{ind,t}}  \right). \label{formula: excess_ret}
\end{equation}

\begin{figure}[ht!]
\centering 
\includegraphics[width=0.9\textwidth]{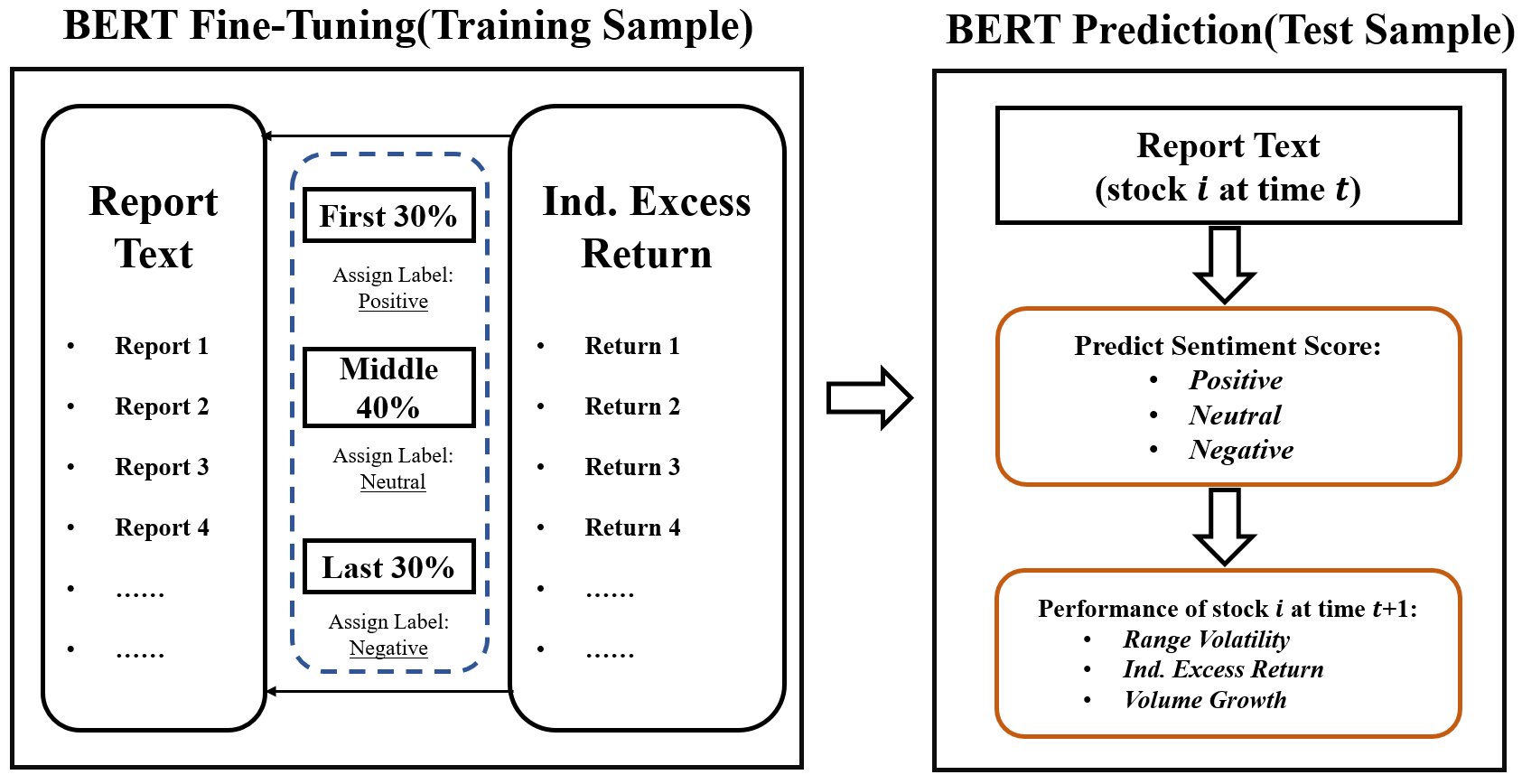}
\caption{Labeling and Prediction of Sentiment}
\label{fig:sentiment_labelling}
\end{figure}

Given the average industrial excess return for each stock, we choose three possible labels: $Pos_{i,t}$, $Neutral_{i,t}$, $Neg_{i,t}$. Labeling is obtained by pooling all excess returns in the eq. (\ref{formula: excess_ret}) and ranking them. If an excess return ranks at the top 30\% among all samples, we consider the corresponding text for the stock $i$ and released on day $t$ to express a positive outlook ($Pos_{i,t} = 1$). Similarly, if the excess return is in the bottom 30\% among all samples, we consider the corresponding text expressing a negative outlook ($Neg_{i,t} = 1$), and if the excess return is placed in the middle 40\%, then we label it as reports of neutral text. Figure \ref{fig:sentiment_labelling} illustrates the workflow for assigning the class label to the training data and the prediction of the sentiment scores based on the test data.

The rationale for this automated labeling is that the analyst in the report observes the immediate performance of the stock compared to its competitors in the industry and may have insight into future returns that are expressed in the opinion piece. Moreover, because all opinion pieces could be biased toward expressing positive sentiments, we adjust this by ranking them, and through this, we derive a balanced sample of positive and negative pieces. We consider this sentiment labeling as merely an indicator rather than a precise quantification of the positivity or negativity expressed in the language. This is a task that cannot be accomplished precisely. 

As this automated labeling process is performed on a training dataset, even if within this dataset the language model is forward-looking, the main predictive results of the paper do not suffer from this forward-looking problem as the language model is applied on a test dataset to generate predicted values of sentiment that are positioned into the future relative to the training dataset.

The language model predicts the probability of classification in each category such that the sum of probabilities of the three categories is one. We treat these probabilities as the intensity of sentiment for each text in each of the three categories. For example, a possible outcome of the prediction is  $\widehat{Pos}_{i,t} = 0.8$, $\widehat{Neutral}_{i,t} = 0.1$ and $\widehat{Neg}_{i,t} = 0.1$. We use the two predicted variables $\widehat{Pos}_{i,t}$ and $\widehat{Neg}_{i,t}$ in the subsequent regression analysis.

\subsection{Regression Model}
\label{subsec:regression}

To analyze the effects created by the text publication, we focus mainly on analyzing the impact of sentiment in the research reports on three variables: industrial excess returns, trading volume difference, and intraday volatility following the report’s release date. By focusing on those measures, we assume that the analyst reports may not only have predictive power, but also have a causal effect on the stock performance. The contents and remarks contained in those research reports can alter the perception of the market.

Further, we focus on the day-after effect of the sentiment in the analyst's report to the stock performance, considering the fact that a liquid and efficient market will promptly react to new information. However, this does not rule out longer-term effects, so we control for past values of performance metrics to capture memory effects as in an auto-regressive process.

The three performance metrics are defined as follows. The industrial excess returns of stock $i$ in day $t$ is:
\begin{equation}
Ret_{i,t}^{ex} = Ret_{i,t} - Ret_{ind,t} \label{eq: stock_excess_ret}
\end{equation}
where $Ret_{i,t}$ is the daily log-return of the stock $i$ in day $t$ and $Ret_{ind,t}$ is the log return of the industry index which the stock belongs to. If the research report text is classified as positive, it should provide additional motivation for investors, thereby resulting in the stock generating excess returns the following day. We only consider the following day because information typically has a time-sensitive nature.

The difference of trading volume is: 
\begin{equation}
    \Delta volume_{i,t} = ln\left(\frac{Volume_{i,t}}{\overline{Volume}_{i,(t-60,t-1})}\right) \label{eq:volume}
\end{equation}
where $Volume_{i,t+1}$ is the trading volume of the stock $i$ for the day $t$ and $\overline{Volume}_{i,(t-60,t-1)}$ is the average daily trading volume of the stock over the past 60 days. A positive research report may increase long positions and thus the trading volume, and a negative research report may also increase short positions, thus the trading volume. However, it is important to note that in the Chinese stock market, short selling is subject to strict limitations; therefore, the volume change may have an asymmetric effect on positive and negative reports.

For intraday volatility, we use the range-based volatility measure that is robust to noise as reviewed by \cite{shu2006testing}. This measure was first introduced by \cite{garman1980estimation} as a model based intra-day volatility estimate from high, low, open and close prices of the trading day:
\begin{equation}
    range_{i,t}=0.511(u-d)^2-0.019{c(u+d)-2ud}-0.383c^2.
\end{equation}
The coefficient estimates are the same as the original article of \cite{garman1980estimation} and the variables $u$, $d$, and $c$  have values that are specific for each stock $i$ and day $t$ (the subscript $i,t$ is omitted for simplicity of notation) and are computed according to \cite{zhang2016distillation}:

\begin{align}
    u &=log(p^H_{i,t})-log(p^O_{i,t})\\
    d &=log(p^L_{i,t})-log(p^O_{i,t})\\
    c &=log(p^C_{i,t})-log(p^O_{i,t})
\end{align}
where $p^H_{i,t}$, $p^L_{i,t}$, $p^O_{i,t}$, $p^C_{i,t}$ are the daily highest, lowest, opening and closing stock prices, respectively for each stock. 

Following the first stage of prediction with the language model, we obtain the variables $\widehat{Pos}_{i,t}$ and $\widehat{Neg}_{i,t}$ that represent the probability that the report is optimistic and pessimistic, respectively. We focus on those two measures and drop the third neutral measure since the sum of the three variables is one at any time $t$. Therefore, we run the following pooled regressions:

\begin{align}
\begin{split}
    Range_{i,t} &= \beta_{0} + \beta_{1} \widehat{Pos}_{i,t-1} + \beta_{2} \widehat{Neg}_{i,t-1} + \beta_{3} Range_{i,t-1} + \beta_{4} Ret^{ex}_{i,t-1} \\
    &+ \beta_{5} \Delta volume_{i,t-1} + \Vec{\theta} \Vec{X} + \varepsilon_{i,t-1} \label{eq: reg1} 
\end{split}
\\[1ex]
\begin{split}
    Ret^{ex}_{i,t} &= \beta_{0}+ \beta_{1} \widehat{Pos}_{i,t-1} + \beta_{2} \widehat{Neg}_{i,t-1} + \beta_{3} Range_{i,t-1} + \beta_{4} Ret^{ex}_{i,t-1} \\
    &+ \beta_{5} \Delta volume_{i,t-1} + \Vec{\theta} \Vec{X} + \varepsilon_{i,t-1} \label{eq: reg2}
\end{split}
\\[1ex]
\begin{split}
    \Delta volume_{i,t} &= \beta_{0}+\beta_{1} \widehat{Pos}_{i,t-1} + \beta_{2} \widehat{Neg}_{i,t-1} + \beta_{3} Range_{i,t-1} + \beta_{4} Ret^{ex}_{i,t-1} \\
    &+ \beta_{5} \Delta volume_{i,t-1} + \Vec{\theta} \Vec{X} + \varepsilon_{i,t-1} \label{eq: reg3}
\end{split}
\end{align}
The parameters $\beta_{0}, \beta_{1}, \beta_{2}, \beta_{3}, \beta_{4}$ and $\theta$ take different values among the regressions (\ref{eq: reg1}) - (\ref{eq: reg3}) (for simplicity we keep the same notation) and the $\varepsilon_{i,t-1}$ are independently distributed.

The vector of parameters $\Vec{\theta}$ is associated with control variables $\Vec{X}$, given by the changes in the VIX index and the returns of three key indices in the Chinese market and the number of times the company $i$ has been cited in the text corpus in previous periods. We use these variables to control the influence of investor sentiment, Chinese market conditions, and firm-specific factors on the regression results.

In summary, with the regression analysis, we focus on three main questions:
\begin{itemize}
    \item Does sentiment score significantly affect Range-Based Volatility? In our hypothesis, both positive and negative sentiment scores would lead to an increase in Range-Based Volatility because intense sentiment would attract more attention to the stock, resulting in increased daily volatility.
    \item Does sentiment score significantly impact the Industrial Excess Return of the following day? If the sentiment score is positive, it should signal the release of positive information, increasing investors' confidence, and leading to higher excess returns the next day. Conversely, if the sentiment score is negative, the effect would be the opposite.
    \item Does the sentiment score significantly affect the volume change on the following day? As for stock volatility, positive and negative sentiments would generate pressure for, respectively, long and short positions, thus increasing trading volume. 
\end{itemize}

To address these questions, we tested the significance of beta coefficients in regressions (\ref{eq: reg1}-\ref{eq: reg3}). In addition, we provide robustness analysis in the following sessions.


\section{Results}
\label{sec: results}

\subsection{Data Source and Data Processing}
\label{subsec: data} 

Our research data are sourced from the commercial financial data platform ``WIND'', which is one of the most popular financial databases in mainland China used by practitioners and researchers. (For reference, see \href{https://www.wind.com.cn/portal/en/EDB/index.html}{here}.)  The text corpus is manually collected over an extended period of time by searching for all published analyst reports and saving the abstract of the article, the release date, and the stock code. It consists of all research reports published by the data platform and made available for scholarly use in the period between March 2017 and February 2023. This data set consists of reports from professional analysts and each report piece is an extensive article. These financial analysts are affiliated with various financial institutions, and their articles may be published elsewhere, for example, on the website or on social media pages of the institution the analyst is affiliated with. As we manually collect the data in a text corpus, we are not aware of any other research paper that makes use of this or part of this dataset. 

We take data from 30 March 2017 to 1 March 2022 as our training data for the BERT language model. It consists of a total of $62735$ records. The test set is made up of data related to the period March 2022 and February 2023 and includes a total of $11101$ records. Overall, the $73836$ text articles are relative to 2506 stocks recommended over 5 years. The number of covered stocks is about half of all stocks listed on the Shanghai and Shenzhen stock exchanges. The purpose of dividing the data sample into training and testing sets is two-fold. First, it is used to fine-tune the BERT model with our financial dataset. This is critical, as the vocabulary and weights in the pre-trained Chinese BERT model are not specifically fine-tuned for financial reports. Second, the impact of text sentiment on stock performance is assessed in the test data set. Because within the training dataset, the sentiment labeling process is forward-looking, we use the test dataset that is positioned into the future to evaluate the impact of the BERT sentiment prediction on stock performance and liquidity, thus avoiding endogeneity problems that could have arisen with the labeling of sentiment based on excess returns.

We process the text data by extracting relevant fields and complementing them with additional information. Furthermore, we performed a pre-processing on the text, by removing irrelevant symbols and spaces. Also, because many reports contain standardized financial risk warnings at the end of the research abstract (these risk warnings often do not represent the analyst's viewpoint, and the inclusion of them is due to industry compliance), we also removed this part of the information that could affect the quality of the text. An example of cleaned text data is presented in Table \ref{tab: data_example}. The title and abstract of the financial report is included as input to the BERT model. In the example, the report text is relative to a specific stock $i$ at time $t$. Also, despite the analyst concluding giving a rating of "outperformance" at the end of his report, this text is labeled as negative sentiment because in the 3 days around the release day the industrial excess return of this stock is negative and ranked low compared to the whole pool of industry excess returns.

\begin{table}[ht]
\caption{An Example of Analyst Report}
\begin{tabular}{p{2cm} p{14cm}}
\hline
Variable & Value \\ 
\hline
Title & {\begin{CJK*}{UTF8}{gkai}2018年盈利同比增长27\%，煤电在建项目持续推进\end{CJK*}}  \\ 
Abstract & {\begin{CJK*}{UTF8}{gkai}公司2018年业绩同比增长27.4\%，18Q4亏损主要受资产减值等影响公司2018年实现归母净利6.6亿元，同比增长27.4\%，折合每股收益0.92元。分季度来看，2018Q1-4分别实现归母净利3.4、2.6、1.5和-1.0亿元，业绩季度环比下降主要是2018年下半年以来成本提升较快(2018Q1-4营业成本分别为10.6, 10.4, 12.8和13.5亿元),近几年公司煤炭业务盈利快速恢复至2011-2012年水平，但电解铝和电力业务盈利下滑，目前公司已针对这部分资产计提减值，后续对公司经营业绩影响有望减小。而通过上大压小，电力业务也有望逐步减亏。中长期来看，公司煤炭在建产能近420万吨，相当于目前在产产能的一半，成长空间也较大。预计2019-2021年公司EPS分别为1.29元、1.34元和1.39元，目前PB仅约0.8倍(行业平均PB约1.2倍，而公司历史平均PB约2.3倍)，公司目前估值较低，认为公司PB估值恢复至1.0倍相对合理，基于2018年底每股净资产，测算公司合理价值为13.2元，维持“增持”评级\end{CJK*}}\\ 
Abstract \\ (translated EN) & The company achieved a year-over-year growth of 27.4\% in its performance in 2018. However, it suffered losses in Q4 2018 mainly due to asset impairment. The net profit attributable to the parent company reached 660 million yuan in 2018, representing an increase of 27.4\% year on year, with an earnings per share of 0.92 yuan. Quarterly, the net profit attributable to the parent company for Q1-Q4 2018 was 340 million yuan, 260 million yuan, 150 million yuan and -100 million yuan, respectively. The decline in quarterly performance was mainly due to the rapid increase in costs since the second half of 2018. In the medium to long term, the company has a coal production capacity under construction of nearly 4.2 million tons, equivalent to half of its current production capacity, indicating significant growth potential. The estimated earnings per share for the company in 2019-2021 are 1.29 yuan, 1.34 yuan, and 1.39 yuan, respectively. Currently, the price-to-book ratio (PB) is only about 0.8 times (compared to the industry average PB of about 1.2 times, and the company's historical average PB of about 2.3 times). Based on the net asset value per share at the end of 2018, the estimated fair value of the company is 13.2 yuan, and an "outperformance" rating is maintained.\\
Stock Code & 600508.SH \\ 
Release & 2019-03-18\\ 
\hline
\end{tabular}
\label{tab: data_example}
    \begin{tablenotes}
    \item Note: The table contains a sample entry from our analyst report dataset with a translation in English provided as reference. The text data has been preprocessed by removing non-text symbols and standardized risk warnings. The relevant variables in this data set are the title, abstract, stock code, and date of publication of the report.
    \end{tablenotes}
\end{table}

For each financial report, corresponding to the pairs $(i,t)$, we complement with additional data as summarized in Table \ref{tab: data_entries}. We download and merge information related to stock prices, stock trading volume, and industrial returns, and further obtain information related to the return of the Shanghai Composite Index ($SSE_{t-1}$), the return of the Shenzhen Composite Index ($SZSE_{t-1}$), the return of the Shanghai High Cap Index ($CSI500_{t-1}$) and the change in the VIX index ($VIX_{t-1}$). Furthermore, based on the reporting time, we calculate the corresponding industrial excess returns ($Ret^{ex}_{i,t-1}$, $Ret^{ex}_{i,t}$), changes in trading volume ($\Delta volume_{i,t-1}$, $\Delta volume_{i,t}$) and range-based volatility ($Range_{i,t-1}$, $Range_{i,t}$) described in Section \ref{sec: methodology}. Finally, we count the number of recommendations that a stock has received in our dataset in the previous three months ($Num_{i,(t-90,t-1)}$) and the previous week ($Num_{i,(t-7,t-1)}$).

\begin{table}[ht]

\caption{Data Entries}
\begin{tabular}{l p{12cm} }
    \hline
    Variable &  Description \\
    \hline
    Report Title & Textual title, typically between 10-30 words \\
    Report Abstract & Main content of the research report \\
    Stock Code & Stock exchange code for stock $i$\\
    Release & The day $t$ the article is published\\
    Price Information & Open price, close, max price, and min price relative to stock $i$ 
 and day to stock $i$ in $t-1$, $t$ and $t+1$ \\
    Stock $Volume$ & Trading Volume relative to stock $i$ and relative to time period $(t-60,t)$\\
    $Ret_{ind}$ & Returns of the industry index to which the stock $i$ belongs at time $t-1$, $t$, and $t+1$\\
    $Ret$ & Continuously compounded return relative to stock $i$ and calculated from closing prices for days $t-1$ and $t$\\
    $\Delta volume$ & Change in trading volume for the stock $i$ relative to time $t-1$ and $t$\\
    $Range$ & Daily estimate of volatility for stock $i$ at time $t-1$ and $t$ and estimated from open price, close price, maximum price, and minimum price.\\
    $SSE$ Market Return & Returns of the Shanghai composite market index at time $t-1$ \\
    $SZSE$ Market Return & Returns of the Shenzen composite market index at time $t-1$ \\
    $SCI500$ Market Return & Returns of the CSI500 (Shanghai top 500 high-capitalization market index) at time $t-1$ \\
    $VIX$ & Implied market volatility index of the CBOT traded options, relative to $t-1$ \\
    $Num7$ & Number of times recommended between $t-7$ and $t-1$ \\
    $Num90$ & Number of times recommended between $t-90$ and $t-1$ \\
    \hline
    \end{tabular}
    \label{tab: data_entries}
    \begin{tablenotes}
    \item Note: This table categorizes all the variables used in our analysis. The table contains four sections. The first section describes the fields extracted from analyst reports texts. The second field describes the complementing data relative to each stock. The third field describes performance metrics relative to stocks, and the fourth field describes market information and control variables such as return from stock indices, VIX index, and stock recommendation counts over specified periods.
    \end{tablenotes}
\end{table}

\subsection{Sentiment Analysis}
\label{subsec: sentiment}

\begin{figure}[H]
\includegraphics[width=1\textwidth]{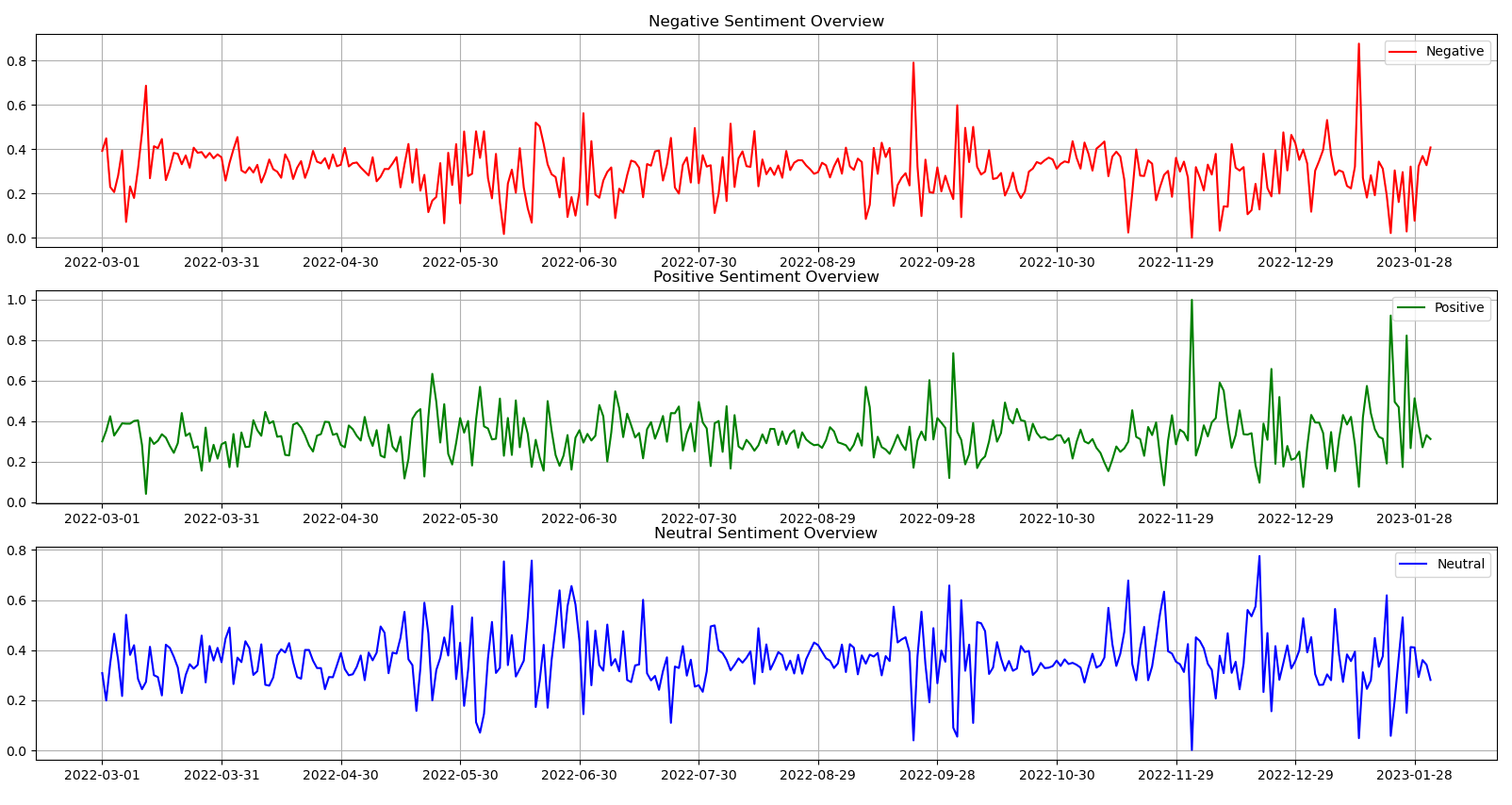}
\caption{Predicted Average Daily Sentiment Score}
\label{fig: sent_score1}
\end{figure}

We present the sentiment scores predicted by the NLP model. Given that the sentiment in the training sample is labeled in a balanced proportion between positive, negative, and neutral, since they are based on the excess return ranking, we expect that the average sentiment for all stocks in the test sample is also balanced between positive, negative, and neutral scores. In Figure \ref{fig: sent_score1}, the average sentiment scores among all stocks are shown for each day for the test sample relative to the period from March 2022 to February 2023. On each day, the sum of positive, neutral, and negative scores is equal to one. Thus, this average score can be interpreted as the market sentiment as conveyed by the analyst reports. 

Overall, we observe a general trend of the average sentiment score fluctuating around 0.3 and 0.4 for all classes throughout the year. Because we provided balanced label classes, the predicted market sentiment is also not significantly more pronounced toward a specific class. However, on certain days, emotions can be particularly high or low, reaching peaks of 0.8 and nadirs of 0. For example, negative sentiment was particularly high on certain days of September 2022 coinciding with a general market decline, and on late December 2022 and early January 2023, coinciding with widespread mass COVID infection in mainland China
On the other hand, in late November 2022, China's relaxation of the ``zero COVID`` policy may have caused a spike in average positive sentiment.

\subsection{Regression Results}
\label{subsec: reg_results}

We present the main results of the regression analysis. The sentiment of analyst reports estimated with the language model is statistically and economically significant for the day after market performance measured as daily volatility, daily industrial excess return, and daily trading volume of the stock. Thus, our result confirms the objective of this study. The results of the regression are reported in Table \ref{tab: Regression_Results}.

\begin{table}[ht]
    \caption{Regression Results}
    \label{tab: Regression_Results}
    \begin{tabular}{p{4cm}p{4cm}p{4cm}p{4cm}}
    \hline
        Variable & Regression of $Range$ & Regression of $Ret^{ex}$ & Regression of $\Delta volume$\\ 
    \hline
        constant & 0.022*** & -0.018 & 0.109*** \\ 
        ~ & (7.482) & (-0.193) & (8.347) \\ 
        $\widehat{Pos}_{t-1}$ & 0.065*** & 0.064*** & 0.177*** \\ 
        ~ & (17.915) & (5.487) & (10.935) \\ 
        $\widehat{Neg}_{t-1}$ & 0.044*** & -0.065*** & 0.036* \\ 
        ~ & (11.537) & (-5.248) & (2.138) \\ 
        $Range_{t-1}$ & 0.291*** & 0.844***& -0.464*** \\ 
        ~ & (48.369) & (4.358) & (-17.337) \\ 
        $\Delta Volume_{t-1}$& 0.002 & 0.064 & 0.539*** \\ 
        ~ & (1.692) & (1.505) & (91.008) \\ 
        $Ret^{ex}_{t-1}$ & -0.004*** & -0.108*** & -0.016*** \\ 
        ~ & (-8.291) & (-6.481) & (-6.997) \\ 
        $SZSE_{t-1}$ & 0.006* & -0.015 & 0.007 \\ 
        ~ & (2.23) & (-0.16) & (0.589) \\ 
        $SSE_{t-1}$ & -0.002 & 0.149* & 0.005 \\ 
        ~ & (-1.365) & (2.547) & (0.575) \\ 
        $CSI500_{t-1}$ & 0.002 & -0.039 & 0.009 \\ 
        ~ & (1.557) & (-1.003) & (1.769) \\ 
        $VIX_{t-1}$ & 0.0007*** & 0.0136*** & -0.0012** \\ 
        ~ & (7.549) & (4.648) & (-3.009) \\ 
        $Num_{t-90, t-1}$ & 0.0033*** & 0.028 & -0.0033 \\ 
        ~ & (3.615) & (0.958) & (-0.834) \\ 
        $Num_{t-7, t-1}$ & -0.0037 & -0.113 & 0.0025 \\ 
        ~ & (-1.885) & (-1.811) & (0.29) \\ 
    \hline
    \end{tabular}

    \begin{tablenotes}
    \item Note: $Range$ is a measure of daily return variance, it has been re-scaled by multiplying by 100. $Ret^{ex}$ is the daily industrial excess return, measured in decimal value rather than percentage. $\Delta volume$ is the ratio between the daily volume of trade and its daily moving average and is measured in decimal value rather than percentage change. $Num_{t-90, t-1}$ and $Num_{t-7, t-1}$ are the counts of the number of citations of the stocks and it measured as hundreads of units (rescaled a unit by dividing by 100)
    \item In parentheses the t-statistics for the test of significance of the coefficients\\
    \item * p-value $< 0.05$, ** p-value $< 0.01$, *** p $< 0.001$ \\
    \end{tablenotes}
\end{table}

First, in terms of range-based volatility, the results confirm the hypothesis that both positive and negative sentiment scores lead to an increase in next day volatility. This result is highly statistically significant. This volatility measure represents intraday fluctuations in stock prices and, in terms of scale, is comparable with daily stock variance. A unit impact of positive sentiment of $0.065$ and a unit impact of negative sentiment if $0.044$ are highly significant economically, since this magnitude represents approximately 2-4 times the average value of this measure. Stock volatility may increase for many reasons. We control for some of those reasons, such as the number of citations the stock received in the past and the volatility memory that is highly persistent, as represented by the estimated coefficient associated with $Range_{t-1}$

The results of range-based volatility infer an asymmetric effect between positive sentiment and negative sentiment. Positive sentiment has a greater magnitude of effect on future volatility than negative sentiment, as the value of the coefficient associated with $Pos_{t-1}$ is approximately double the value of the coefficient associated with $Neg_{t-1}$. This result seems to contradict the traditional volatility asymmetry effect such as that presented by early studies such as \cite{bekaert2000asymmetric}, in which negative news has a greater effect on stock volatility compared to positive news. Instead, this result is consistent with the "reverse" volatility asymmetry effect presented by \cite{wan2014reverse} for the Chinese market, in which positive returns have a greater effect with respect to negative returns to future volatility. We also believe that due to short-sale restrictions in the Chinese market, the market dynamics due to negative news could be less pronounced compared to those due to positive news.

Our second hypothesis is about the impact of sentiment on the excess returns on the next day. Reports with positive sentiment should increase investors' confidence, thus leading to higher excess returns, and, conversely, reports with negative sentiment should generate the opposite effect. The regression results of $Ret^{ex}$ confirm this hypothesis. The coefficient associated with $\widehat{Pos}_{t-1}$ and $\widehat{Neg}_{t-1}$ is highly statistically and economically significant. A unit of estimated sentiment has a partial effect in the magnitude of about $0.064$ on the daily excess return rate. As the sentiment score ranges between 0 and 1 this means that absolutely positive reports have an estimated $6.4\%$ impact on future excess returns, while absolutely negative reports have about the opposite effect of $-6.5\%$. This represents an extreme as absolutely positive or negative sentiment is rarely estimated in the data, and the linear regression model weights more on the most frequent values for the sentiment score, which is about 0.3 for both positive and negative sentiments.  

Our third hypothesis is about the impact of sentiment on the next-day change in volume. Similarly, as for stock volatility, both positive and negative sentiment reports should increase stock volume due to increased pressure from long and short positions. We find that, while positive sentiment has a statistically and economically significant effect, the effect of negative sentiment on volume change is much less significant. In the regression of $\Delta volume$, the coefficient associated with $\widehat{Neg}_{t-1}$ compared to that associated with $\widehat{Pos}_{t-1}$ is negligible. The significance test for the former coefficient does not pass at the level $0.01$. We interpret this asymmetric effect as the asymmetric effect on the volatility of the stock, as studied in \cite{wan2014reverse}. Also, in the Chinese market only certain professional investors can short-sell and with controlled volume limits. Therefore, we believe that these short-selling restrictions may lead to a less pronounced effect for negative-sentiment reports. It is also possible that negative sentiment has less of an impact on future volume compared to positive sentiment due to behavioral and socioeconomic factors. Retail investors make up the majority of the market, and China has seen an extraordinary increase in private wealth, leading to a dominant sentiment of optimism and excitement among these investors. As a result, Chinese investors are less sensitive to negative news than to positive news. We reserve this argument for further research.

\subsection{Robustness Check: Performance by Industry}
\label{subsec: robust}

As a robustness check, we split the entire dataset into 16 subsets based on industries and perform the same regressions (\ref{eq: reg1}) - (\ref{eq: reg3}) on each of these subsets. This allows us to identify the impact of potential outliers, such as specific stocks that exhibit a strong sensitivity to estimated sentiment and may be driving the overall results. When industry regressions are run, only the results of the respective industry would be affected by these extreme values. 

\begin{table}[ht]
    \caption{Industry Distribution of Stocks in the Test Set}
    \label{tab: industry_distribution}
    \begin{threeparttable}
    \begin{tabular}{p{5cm}p{5cm}p{5cm}}
    \hline
        Industry & Sample Size & Num. Unique Stocks \\
    \hline
        Material & 1765 & 311 \\
        Telecom & 1422 & 251 \\
        Real Estate & 218 & 26 \\
        Public utilities & 372 & 41 \\
        Media & 174 & 26 \\
        Apparel & 842 & 98 \\
        Automobile & 422 & 59 \\
        Business Service & 302 & 46 \\
        Food, staples, retail & 1200 & 104 \\
        Consumer Service & 173 & 17\\
        Healthcare & 316 & 41\\
        Bank & 365 & 36\\
        Transport & 270 & 45\\
        BioTech & 676 & 108\\
        Capital Goods & 1446 & 283\\
        Other & 1134 & 168\\
    \hline  
    \end{tabular}
    \begin{tablenotes}
     \item Note: This table shows the industry distribution of stocks in our test set. Stocks without multiple classifications and minor sector categories are classified as "Other".
    \end{tablenotes}
    \end{threeparttable}
\end{table}

Furthermore, by examining the effect of sentiment on stock performance within each industry, we investigate industry-specific effects that are important in the Chinese market. Certain sectors, such as the banking industry, have a high concentration of Chinese state-owned companies. We hypothesize that these sectors are less responsive to sentiment effects due to the significant role government control plays in stabilizing market performance.

The results of the regression analysis of the performance of the sentiment of the stock for each of the 15 industries sectors are presented in Table \ref{tab: regression_industry}. Only the coefficient associated with $\widehat{Pos}_{t-1}$ and with $\widehat{Neg}_{t-1}$ are reported for each of the performance metrics $Range$, $Ret^{ex}$ and $\Delta volume$.

\begin{table}[ht]
    \caption{Regression on Industry Subsets}
    \label{tab: regression_industry}
    \begin{threeparttable}
    \resizebox{\textwidth}{!}{
    \begin{tabular}{p{3cm} p{1.5cm} p{1.5cm} p{1.5cm} p{1.5cm} p{1.5cm} p{1.5cm}}
    \hline
         ~  & \multicolumn{2}{c}{Regression of $Range$}  & \multicolumn{2}{c}{Regression of  $Ret^{ex}$} & \multicolumn{2}{c}{Regression of 
 $\Delta volume$} \\ 
        Industry & $\widehat{Pos}_{t-1}$ & $\widehat{Neg}_{t-1}$ & $\widehat{Pos}_{t-1}$ & $\widehat{Neg}_{t-1}$ & $\widehat{Pos}_{t-1}$ & $\widehat{Neg}_{t-1}$\\ 
        \hline
        Material & 0.033* & 0.007 & 0.283 & -0.924 & 0.006 & -0.033\\ 
        ~ & (1.947) & (0.3804) & (0.454) & (-1.375) & (0.093) & (-0.4503)  \\ 
        Telecom & 0.022 & 0.007 & 1.145** & -0.183 & 0.064 & -0.0197  \\ 
        ~ & (1.147) & (0.382) & (1.995) & (-0.313) & (1.016) & (-0.308)  \\ 
        Real Estate & 0.018 & 0.009 & 2.289* & -1.669 & -0.009 & -0.107  \\ 
        ~ & (0.529) & (0.221) & (1.729) & (-1.044) & (-0.058) & (-0.582)  \\ 
        Public Utilities & -0.005 & 0.015 & 0.167 & 0.014 & 0.028 & 0.015  \\ 
        ~ & (-0.216) & (0.769) & (0.192) & (0.017) & (0.248) & (0.142)  \\ 
        Media & 0.046 & 0.094** & -2.822 & -0.644 & -0.174 & 0.161  \\ 
        ~ & (1.034) & (2.032) & (-1.555) & (-0.343) & (-0.923) & (0.828)  \\ 
        Apparel & 0.0702*** & 0.036 & 1.775** & 0.591 & 0.136 & -0.004  \\ 
        ~ & (2.749) & (1.367) & (2.124) & (0.692) & (1.374) & (-0.036)  \\ 
        Automobile & 0.107*** & 0.098** & 2.292** & 0.994 & 0.185 & 0.087  \\ 
        ~ & (3.003) & (2.313) & (1.962) & (0.718) & (1.553) & (0.617)  \\ 
        Business Service & 0.011 & 0.031 & 1.899 & 1.224 & 0.036 & -0.097  \\ 
        ~ & (0.258) & (0.755) & (1.263) & (0.809) & (0.211) & (-0.569)  \\ 
        Food, staples, retail & 0.017 & 0.017 & 0.714 & -0.470 & 0.161** & 0.139*  \\ 
        ~ & (0.802) & (0.807) & (1.088) & (-0.729) & (2.089) & (1.835)  \\ 
        Consumer Service & 0.032 & -0.041 & 1.454 & 1.733 & 0.045 & 0.017  \\ 
        ~ & (0.533) & (-0.583) & (0.726) & (0.733) & (0.237) & (0.073)  \\ 
        Healthcare & 0.192** & 0.182** & -0.871 & -2.185 & 0.254 & 0.037  \\ 
        ~ & (2.175) & (2.016) & (-0.413) & (-1.011) & (0.996) & (0.141)  \\ 
        Bank & 0.017*** & 0.021*** & 0.048 & 0.248 & 0.073 & -0.007  \\ 
        ~ & (3.095) & (3.125) & (0.127) & (0.516) & (0.823) & (-0.0603)  \\ 
        Transport & -0.059* & -0.049 & -1.068 & -2.278* & -0.176 & -0.004  \\ 
        ~ & (-1.908) & (-1.442) & (-0.889) & (-1.757) & (-1.462) & (-0.034)  \\ 
        BioTech & -0.027 & -0.005 & -0.769 & -1.8403 & 0.032 & 0.048  \\ 
        ~ & (-0.679) & (-0.122) & (-0.554) & (-1.352) & (0.216) & (0.329)  \\ 
        Capital Goods & 0.027 & 0.033 & 0.468 & -0.025 & 0.029 & -0.017  \\ 
        ~ & (1.404) & (1.556) & (0.721) & (-0.036) & (0.391) & (-0.209) \\ 
    \hline
    \end{tabular}}

    \begin{tablenotes}
    \item Note: $Range$ is a measure of the daily return variance, it has been rescaled by multiplying by 100. $Ret^{ex}$ is the daily industrial excess return, it is measured in decimal value rather than percentage. $\Delta volume$ is the ratio between the daily volume of trading and its daily moving average and is measured in decimal value rather than percentage change. 
    \item The t statistics are given in parentheses.
    \item * p-value $< 0.1$, ** p-value$< 0.05$, *** p-value $< 0.01$.
    \end{tablenotes}
    \end{threeparttable}
    
\end{table}

The results of this analysis are mixed. In general, we confirm the main empirical findings of significant impact of the sentiment of the report on the performance metrics of the stock market, but with some exceptions. The range-based volatility analysis ($Range$) is the most robust, with seven industry subsets producing significant results, which consist of our previous analysis. For the industrial excess return ($Ret^{ex}$), five subsets of industries show significant results according to the main results. Finally, for the difference in the trading volume ($\Delta volume$), only one subset of industries exhibits weakly significant results according to the main predictions. 

The lack of significant results may be influenced by the smaller size of the industry data set. The whole test dataset has about 11 thousand text reports, but for some industries only a few hundreads of citations are present and are relative to a dozen stocks each. For example, ``Consumer Service'' and ``Media'' are sectors with the smallest sample size with insignificant results. Table \ref{tab: industry_distribution} presents the distribution of analyst reports on stocks by different industries. There are sectors with a relatively high number of recommendations, such as ``BioTech'' and ``Capital Goods'', but with a not significant sensitivity to market sentiment. An explanation for this is the hypothesis that some sectors are less sensitive to market news because of government control.

On the contradictory side, we have the results in the industries ``Transport`` that have an opposite coefficient sign, specifically for the sentiment effect in regressions on $Range$. The main reason could be the limit of the size of the subset. Since we only have 270 samples in this industry, and a few outliers are driving opposite results.

The asymmetry in (range) volatility to sentiment and volume change to sentiment is confirmed in this industry-based analysis. Although some of the results are not statistically significant, for significant industries, this effect persists. 

\subsection{Robustness Check: Manual Labeling and Statistical Tests}
\label{subsec:stat_test}
To further validate our main results, we run statistical tests for the difference in means in a manually labeled subset of data. We use the Chinese text segmentation module "Jieba"\footnote{A Python Chinese word segmentation module: https://github.com/fxsjy/jieba} to extract all Chinese worlds from the research reports. We manually annotated the top 1000 words with the highest frequency as positive, negative, or neutral. Some sample annotations with the corresponding English translation are listed in table\ref{tab: Keywords}.

Based on these manually labeled keywords, we categorized the reports into three classes: positive if there were more positive words than negative ones, negative if the opposite was true, and neutral if the occurrences of positive and negative words were equal. Subsequently, we performed difference in-mean tests in the positive and negative groups of reports, with the results presented in the table\ref{tab: t_test}.

\begin{table}[ht]
\centering
    \caption{Sample of Word Labeling}
    \label{tab: Keywords}
    \begin{tabular}{cp{5cm} p{8cm}}
        \hline
        ~ & Example Words & English Translation \\
        \hline
        Positive & \begin{CJK*}{UTF8}{gbsn}增持, 快速增长, 突破, 领先地位, 中标, 龙头, 有望, 看好\end{CJK*} & Increase holdings, rapid growth, breakthrough, leading position, win the bid, dragon head, hopeful, be optimistic \\
        Neutral & \begin{CJK*}{UTF8}{gbsn}整体, 逐步, 长期, 调整, 景气, 恢复, 稳步\end{CJK*}  & Overall, step by step, long-term, adjustment, prosperity, recover, steadily  \\
        Negative & \begin{CJK*}{UTF8}{gbsn}疫情, 下滑, 下调, 审慎, 减值, 损失, 不确定性, 竞争\end{CJK*} & Epidemic, decline, downside, careful, impairment, loss, uncertainty, compete\\
        \hline
    \end{tabular}
\end{table}

\begin{table}[ht]
  \centering
    \caption{Test for the Difference in Means}
    \label{tab: t_test}
    \begin{threeparttable}
    
    \begin{tabular}{p{2.5cm} p{2cm} p{2cm} p{2cm} p{2cm} p{2.5cm}}
    \hline
    ~  & \multicolumn{2}{c}{Positive Group}  & \multicolumn{2}{c}{Negative Group} & Mean Difference\\
    ~ & Mean & Standard Deviation & Mean & Standard Deviation & $t$-Stat \\
    \hline
    $Ret^{ex}_{t}$ & 0.767 & 4.222 & 0.112 & 4.128 & 11.931***\\
    $Ret^{ex}_{t-1}$ & 0.454 & 3.639 & 0.058 & 3.447 & 8.561***\\
    $Ret^{ex}_{t+1}$ & 0.344 & 3.546 & 0.189 & 3.427 & 3.382***\\
    $Ret^{ex}_{3DayAverage}$ & 0.522 & 2.294 & 0.119 & 2.228 & 13.539***\\
    $\Delta volume$ & 0.131 & 0.564 & 0.147 & 0.553 & -2.123*\\
    $Range$ & 0.119 & 0.131 & 0.105 & 0.123 & 8.490***\\
    \hline
    \end{tabular}

    \begin{tablenotes}
    \item Note: $Ret^{ex}_{t\pm1}$ is the daily industrial excess return after/before the release date, it is measured in decimal value rather than percentage. $Ret^{ex}_{3DayAverage}$ is the average industrial excess return of 3 days around the release date.
    \item * p-value $< 0.1$, ** p-value$< 0.05$, *** p-value $< 0.01$.
    \end{tablenotes}
    \end{threeparttable}
    
\end{table}

The results of the statistical tests are largely consistent with our previous analysis. The group labeled as positive, influenced by the release of research reports, exhibits higher industry excess returns.

\section{Conclusion}
\label{sec: conclusion}

In this paper, we collect an extensive dataset of institutional analyst stock recommendations that spans five years and use this data set to train a language model based on the Chinese BERT model of \cite{cui-etal-2021-pretrain}. We use this model to predict the sentiment conveyed in the reports. Based on this sentiment measure, we analyze their impact on stock performance as excess return, volatility, and trading volume. 

The results show that the sentiment of the report has a significant impact on the performance of the stock. Stock volatility and trading volume increase after positive and negative sentiment, with negative sentiment impacting less than positive sentiment. The excess return tends to increase after positive sentiment reports and tends to decrease after negative sentiment reports. These results are consistent with previous literature on sentiment pricing and news distillation (\cite{zhang2016distillation}). Unlike the early established points, we find that the asymmetric effect of news sentiment on trading volume and stock volatility is a ``reverse`` asymmetric effect for the Chinese market, similar to what has been found in \cite{wan2014reverse}, where positive news has more of an impact with respect to negative news. 

We performed several robustness checks on our results. First, by running regressions on industry data subsets and then by testing for difference in mean return for a subset of data based on count of manually labeled positive and negative worlds. Despite the fact that for some of the industries the main result is not statistically significant, we retain the overall results on abnormal excess returns. 

For further research, the Chinese stock market has unique characteristics that make it an interesting subject of study. For example, government influence on the market dynamics by means of corporate control and governance of, for example, critical economic sectors. Given this balanced dual participation of private and state in the market economy, we expect that there is a large degree of heterogeneity in the stock price dynamics, especially as a response to news.

Our work contributes to the empirical literature on the Chinese stock market and sentiment analysis. We find great potential in applying state-of-the-art deep learning models to Chinese text reports. To our knowledge, there is no extensive prior research applying Chinese language models to financial texts and using this for stock evaluation. Our paper fills this gap.


\newpage

\bibliography{reference.bib}




\end{document}